# Applying Evolutionary Algorithms Successfully
## A Guide Gained from Real-world Applications

Wilfried Jakob

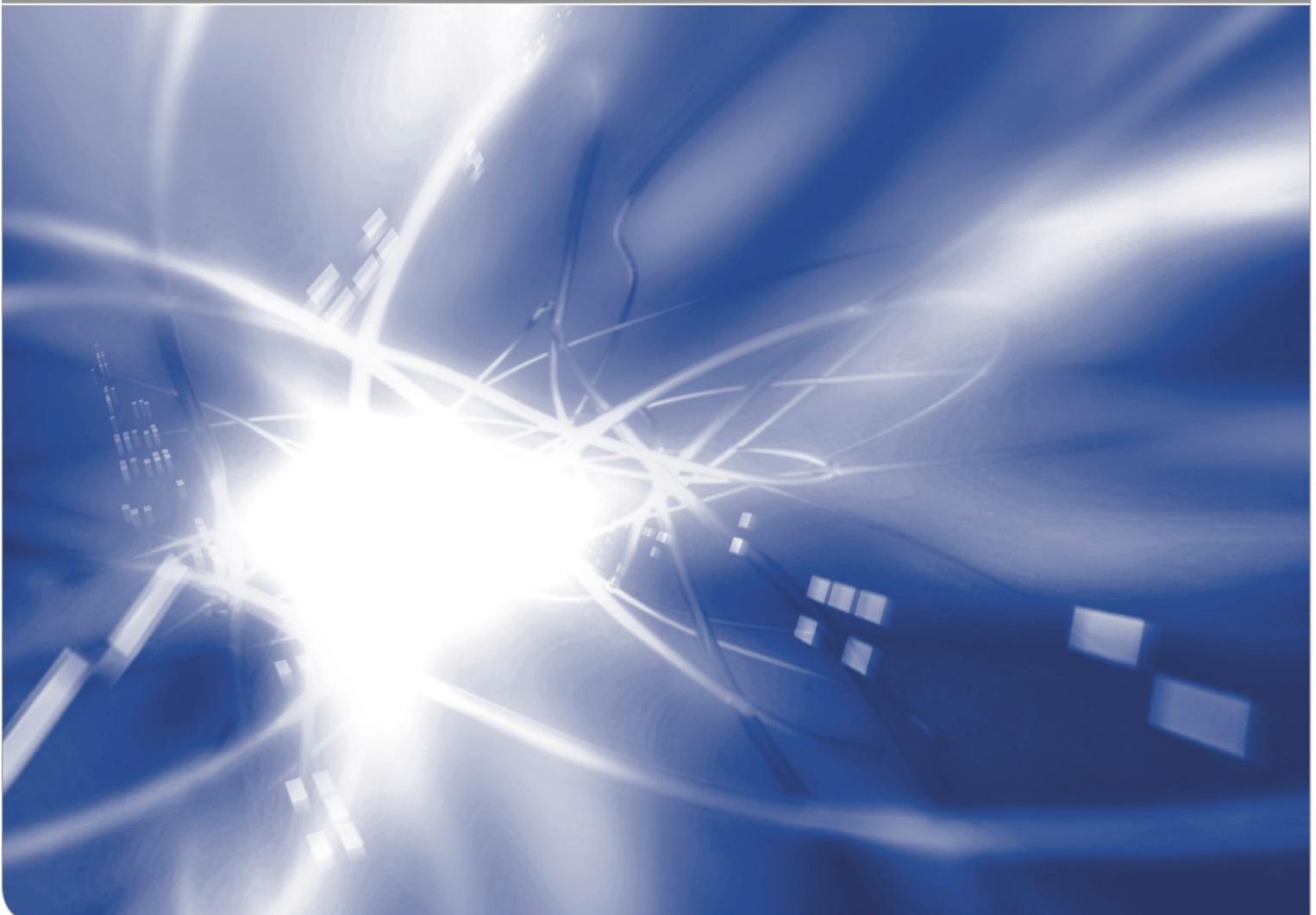

**Institute for Automation and Applied Informatics (IAI)**

**Impressum**

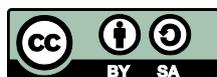

# Applying Evolutionary Algorithms Successfully
## A Guide Gained from Real-world Applications

## Abstract

Metaheuristics (MHs) in general and Evolutionary Algorithms (EAs) in particular are well known tools for successful optimization of difficult problems. But when is their application meaningful and how does one approach such a project as a novice? How do you avoid beginner's mistakes or use the design possibilities of a metaheuristic search as efficiently as possible? This paper tries to give answers to these questions based on 30 years of research and application of the Evolutionary Algorithm GLEAM and its memetic extension HyGLEAM. Most of the experience gathered and discussed here can also be applied to the use of other metaheuristics such as ant algorithms or particle swarm optimization.

This paper addresses users with basic knowledge of MHs in general and EAs in particular who want to apply them in an optimization project. For this purpose, a number of questions that arise in the course of such a project are addressed. At the end, some non-technical project management issues are discussed, whose importance for project success is often underestimated.

## Contents

# 1. Introduction

Evolutionary Algorithms (EAs) [1 - 6] reproduce certain mechanisms of biological evolution as algorithms to improve solutions according to given objectives. These mechanisms include the inheritance of information (crossover and recombination) that determines the solution, its random variation (mutations) and the parallel treatment of a set of solutions (population). The chance of a solution (individual) to produce offspring and/or to be present in the next iteration of the algorithm (generation) depends on its quality (fitness).

This general mechanism of generating and (randomly) changing a set of solutions and evolving them based on a quality measure can, with appropriate abstraction, be found in many other metaheuristics (MHs) such as ant colony algorithms [7] or particle swarm optimization [8] as well. Therefore, many of the more general statements made here about EAs also apply to the other methods. The focus of this paper is on population-based, stochastic search methods that globally scan the search space and are independent of a specific application.

# 2. When Should MHs Be Used and When Better Not?

If there is an exact mathematical solution to a problem and one is satisfied with the results, there is usually no reason to use a metaheuristic. If simplifications of the problem were necessary to be able to apply an exact (established) solution method, the question arises, how large are these simplifications and how far are the found approximate solutions from the original problem? This question can only be answered application-related and in case of doubt only the application of a suitable MH and a comparison of the results will help.

Exact mathematical solutions can generally not be expected for the following types of tasks:

- NP-complete problems [9], including most combinatorial tasks, such as the Traveling Salesman Problem (TSP, cf. Section 4.3) or scheduling problems above a certain (usually small) size,
- Tasks whose mathematical objective function is "difficult" because, for example, it is characterized by non-linearities, discontinuities, or regions with definition gaps.

In all these cases, metaheuristic optimization is a promising solution approach. Here are some examples:

- Design optimization, e.g. of turbines, aircraft wings, antennas, or printed circuit board layouts,
- Scheduling tasks, like production planning, timetable generation for schools or public transport, maintenance planning,
- Layout planning, like cutting planning (textile industry, shipbuilding, ...), container or truck loading,
- Site planning based on standard tours and/or other restrictions, power plant or network expansion based on different load scenarios.

In addition, there are all kinds of model-based optimization. The creation of simulation models usually aims at calculating and evaluating alternatives. The variation of parameters can be based on experience and intuition or more comprehensively by a MH. Both can also be complemented by using manually found solutions as starting points of a global search. Using e.g. an elitist EA[1] the result can only get better.

Metaheuristics have advantages and disadvantages. The disadvantages include:

- Stochastic search, so results cannot be reproduced. This means that two runs will usually return different results.
- No guarantee of finding the optimum. But (very) good solutions can be expected.
- Comparatively long run times. It must be possible to compute a large number of alternative solutions.

This is offset by the following advantages:

- General applicability (at least related to a problem class),
- Global search in the solution space,
- Simplifications of the problem are neither necessary nor indicated. The only exception is for tasks requiring long run times for solution assessments. If they are e.g. simulation-based, simplifications can be considered if they have only a small and estimable effect on the results.
- Manually, heuristically or otherwise generated solutions can be integrated as initial individuals,
- Good parallelizability as acceleration measure,
- With faster hardware, better solutions or the possibility of scaling the task higher[2] can be expected with the same runtime. A third alternative is faster processing of the unchanged task with comparable solution quality.
- Good combinability with local search methods to find good solutions more reliably and faster.

Another advantage can be seen in the low prerequisites that are necessary for an EA application:

- Essentially, only adherence to the principle of causality is expected, i.e., that like causes have like effects, and the greater the strength of causality, i.e., the more strongly the magnitude of a cause change is related to the magnitude of the effect change, the better. As is true for almost all other search methods, EAs are unlikely to find the lone peak in an otherwise flat search space. In such a situation, purely random-based approaches such as the Monte Carlo method are most promising. This is because EAs try to learn from the search space samples and if there is nothing to learn, search strategies based on that will not help.
- It must be possible to evaluate solutions in such a way that their quality is comparable.
- Sufficient time and/or computing power is available, whereby it depends on the application what can be considered as "sufficient".

---

1  In an elitist EA, the best individual of a population can only be replaced by a better one.
2  e.g. in the case of a scheduling task by increasing the planning volume

The following properties of quality functions, on the other hand, are not an obstacle to EA application:
- non-linearity,
- only section-wise definition,
- missing differentiability,
- missing continuity,
- missing convexity, or
- noisiness of the function.

## 3. Which Metaheuristic Is Suited?

Evolutionary Algorithms like GLEAM [6, 10, 11], their memetic extensions [12, 13] like HyGLEAM [14, 11], ant colony optimization [7], and particle swarm optimization [8] belong the oldest population based global searching metaheuristics. They are well researched and there are a lot of applications, so it should be easy to find application examples for your own task in the literature. If not, at least reports of similar tasks should be found to learn from.

An important criterion for the selection of a MH is the question of how easily the decision variables and other degrees of freedom of a task, discussed below in Section 4.1, can be mapped to the MH under consideration. For EAs, this would be the issue of easy assignment to genes and chromosomes. Also for this question, which is important for success, one should try to learn from the applications of others.

Next, the characteristics of the MH must fit the type of the task (see Section 4.1.2). In combinatorial problems, when using, for example, an EA, it is useful that its genetic operators enable or, better, foster a change in gene order. Thus, it can be (very) helpful if, in addition to a simple mutation for gene shifting, there are also those that shift entire gene sequences [15] or reverse their internal gene order (inversion) [15]. In addition to suitable mutations, special crossover operators may also be useful, such as order-based crossover [5, 16], which passes the relative gene order to the offspring. This crossover is useful, for example, in scheduling tasks that have sequence constraints. The classification and categorization of an optimization task will be discussed in more detail in Section 4.1.2.

A further criterion for selection is the availability of suitable software. Suitability issues include the support that can be expected and whether the sources are available for any necessary extensions or modifications.

Recently, a variety of novel nature-inspired search methods have been reported [17], such as bee [18], firefly [19] or gray wolf [20] algorithms to name only a few. An advantage over the aforementioned methods is not apparent at first, and as long as these cannot report a comparable depth of research and number of applications as the established MHs, caution is advised for the beginner. Because when problems arise, there is a risk of finding rather little in the literature.

## 4. From the Problem Statement to an EA Project

This section describes how to derive an optimization project from a task definition. This step is explained on the basis of the properties of an EA in order to keep the text simpler and more comprehensible. However, most of the statements can also be applied to other metaheuristics based on a global search. The approach presented is illustrated in Section 4.3 by discussing some practical examples.

First, the following basic preliminary consideration: For EAs, the success of the search is based on two main requirements: First, the information contained in the chromosomes must cover all areas of the search space to be considered. Second, the assessment must be comprehensive in the sense of the objectives. One must always keep in mind that a basic EA has no knowledge of the optimization goals and will therefore proceed strictly according to the user-specified evaluation criteria to steer the evolution. If, for example, a ship is to be planned, it is clear to everyone involved that it must be able to swim. But if this requirement is not part of the evaluation, one should not be surprised about too heavy ship designs.

### 4.1. Objectives, Decision Variables, and Constraints

In order to move from a task definition to an optimization project with an EA, the following elements must first be identified in the task description.

#### 4.1.1. What are the Goals of the Optimization?

Evaluation criteria are deduced from the goals. They are also called *primary criteria*, since they can be derived directly from the task. The topic of evaluation and fitness calculation will be discussed further below in Section 4.4. Depending on the task, it may be useful to define additional criteria that support the fulfillment of the primary ones, see Section 4.2. It should be noted at this point that compliance with restrictions can also be included in the evaluation, see Section 4.1.3.

#### 4.1.2. What are the Changeable Variables

What are the parameters that are to be determined by the optimization? Are there other degrees of freedom that can or must be used to find a solution? The parameters to be set are called *decision variables* (DVs). Further degrees of freedom may concern the order in which solutions are constructed from partial steps. This can be e.g. scheduling operations, where resources are assigned to a sub-operation like a work step. Or in a layout planning operation, an element to be placed is selected and assigned to a location.

The DVs and any sequence information are mapped to genes, which then form the chromosome of an individual. This step is usually referred to as *coding* in the literature. To calculate the fitness of an individual, the reverse step is required and a solution must be constructed from the information contained in the chromosome, which can then be evaluated. This step is also called *interpretation* of a chromosome. In this step, data that are not subject to optimization but are relevant for the construction of a solution (static data) may also be used. Coding and interpretation are usually developed jointly after the identification of the DVs, as they are closely interdependent.

DVs can be integer or continuous and it would be best if they could also become part of a chromosome unchanged with this property. Unfortunately, many if not most EA implementations are based on chromosomes with either real or integer numbers. In such cases, one must either round or restrict real numbers to a suitable precision and map them to integers.

An optimization problem can be classified on the basis of its DV and other degrees of freedom as follows, and this classification can be helpful in selecting and finding a suitable MH.

- Optimization of continuous and/or integer parameters. If both data types occur, it is also called mixed-integer optimization. Example: Design optimization
  Some appropriate MHs: Evolution Strategy [2, 3], real- or integer-coded GAs [4, 21], Particel Swarm Optimization [8], HyGLEAM [14, 11]

- Combinatorial optimization. Examples: TSP (see Section 4.3.1), some scheduling tasks.
  Some appropriate MHs: Ant Colony Optimization [7], Integer-coded GAs [4], GLEAM [6, 10, 11]

- Combinatorial including mixed-integer optimization. Examples: some scheduling tasks (see Section 4.3.2), layout planning (see Section 4.3.3)
  Some appropriate MHs: real- or integer-coded GAs [4, 21], GLEAM [6, 10, 11] or HyGLEAM [15, 11]

The above list of suited MHs is by no means complete and contains mainly those MHs which were designed for this field of application. For the other application classes, there are usually correspondingly adapted variants.

A note on the classical genetic algorithms (GAs), since they are still relatively widespread. They are based on a binary coding and represent everything as bit strings of different lengths. Michalewicz already analyzed the serious disadvantages of this coding for all tasks not based on Boolean DVs and/or integers of small ranges in his book about 30 years ago [4]. Real-coded GAs or those based on integers can be considered as reasonable alternatives to the classical form.

### 4.1.3. Which Restrictions are There?

Real-world applications usually have some restrictions. The author has never seen a real-world problem without them in about 30 years of research and EA application.

The value range of DVs is usually limited. Such restrictions are called *explicit restrictions* and a good EA allows to specify their lower and upper bounds in such a way that the mutations respect them. Restrictions that depend on multiple DVs are called *implicit restrictions* and can be thought of as illegal regions in or at the edge of the search space.

The simplest and also worst way to deal with implicit restrictions is to set the fitness of such an individual to zero. Instead, the degree of restriction violation should be determined and evaluated so that the individual has a chance to move out of the forbidden area[3]. The more it approaches the boundary of the impermissible zone, the better the fitness must become until it increases signifi-cantly once the forbidden zone has been left. Only in this way do initially impermissible solutions have a chance of becoming permissible ones.

---

3  The image used here of an individual moving based on fitness increases is intended to serve as an easier representation of the search process. In reality, of course, a change occurs through improved offspring of the initial individual in the course of generations and since the best offspring replaces its parent in many EAs, the image of the moving and improving individual arises over the generations.

Figure 1 may illustrate this. The left part (a) shows the original fitness function. It is usually not sufficient to simply lower the fitness value in the forbidden zone, because then the peak labeled *P* in the figure will cause individuals to move there rather than to the restriction limits, see Fig. 1b. Instead, use a suitable function in the forbidden region, e.g., a lowered spherical function as in Fig. 1c. This helps the individuals to move to the boundaries and to get a significant fitness increase after leaving the forbidden zone, which considerably counteracts a return.

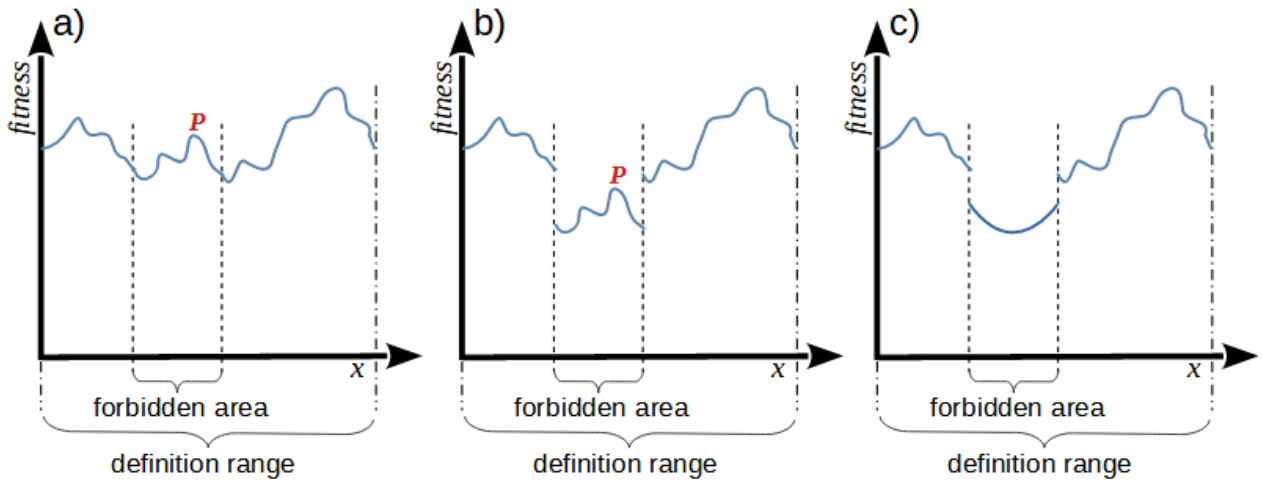

**Fig. 1:** Handling of a restriction within the definition range of a DV. The original fitness function (a), a simple and not sufficient modification (b), and a more promising adjustment (c).

Handling restrictions by evaluation is an option that always exists. Unfortunately, depending on the application, it can also become quite laborious and make it more difficult for the search. The better option is to look for ways to completely avoid violations of at least some of the restrictions by choosing a clever design of chromosomes and their genes, as well as their interpretation. To this end, the examples in Section 4.3 will hopefully provide some ideas.

### 4.1.4. Genotypic and Phenotypic Repair

In a number of applications, it can be determined during interpretation whether certain restrictions are violated by the actual gene or not. This can be responded to by two fundamentally different types of repair, genotypic and phenotypic repair, in addition to the always possible lowering of fitness discussed in Section 4.1.3. In the former, repair is accomplished by appropriately altering the chromosome, whereas in the latter, only the interpretation is adjusted and the chromosome remains unchanged. As a consequence, in the case of phenotypic repair, this must also be applied to the interpretation and output of the final result.

An example of this are all scheduling tasks in which there are sequence specifications for at least some of the sub-steps. When processing a gene, it can now be determined whether all required sub-steps will be completed in time before the scheduling in question. If not, there is a restriction violation that can be easily remedied phenotypically: the processing of the affected gene is simply postponed until all required predecessor sub-steps have been scheduled. In genotypic repair, on the other hand, the affected gene is shifted towards the end of the chromosome until it is behind all genes of the predecessor sub-steps. The disadvantage of this procedure is that it prevents the restructuring of the gene sequence distributed over several appropriately modified offspring, which have restriction violations in between.

## 4.2. Auxiliary Criteria

Auxiliary criteria do not come from the original problem definition, but are intended to support and accelerate the achievement of certain criteria. As an example, consider production planning where one of the goals is to shape the energy demand resulting from the planning in such a way that consumption peaks above a given limit are as low as possible or preferably avoided completely, see Fig. 2a. This results in the primary goals of a low peak count and a low peak maximum. But this is not sufficient, as the following consideration shows: If one of several peaks of the same size is reduced by a suitable change in the schedule, this has no effect on either the number of peaks or their maximum, see Fig. 2b, 1$^{st}$ peak. However, the improvement that nevertheless occurs would have to be evaluated positively, because it is probably a meaningful intermediate step towards a better schedule. By the way, the same applies to the reduction of the temporal duration of a peak, see Fig. 2b, 5$^{th}$ peak. Thus, the area of the peaks above the limit should be calculated and summed up and included in the overall evaluation. In use cases, where schedules without energy peaks are possible, the evaluation of the peak maximum is probably no longer necessary.

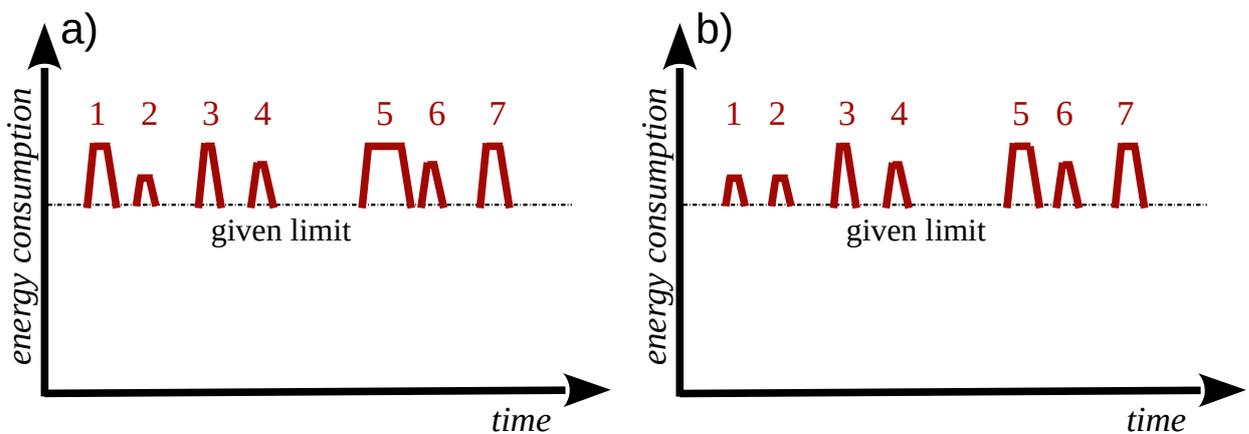

**Fig. 2:** Assessment of energy peaks exceeding a given maximum. The situation in the right part of the figure shows an improvement in the 1$^{st}$ and 5$^{th}$ peak compared to the left one.

The above application example can be used to demonstrate another use of auxiliary criteria. As a rule, compliance with the latest completion times of the orders represents an important evaluation criterion. In order to shorten this if necessary, all processing steps of the order must start as early as possible and delays between the steps must be kept as short as possible. Therefore, simply evaluating compliance with the deadlines is not sufficient; instead, the waiting times of the individual processing steps should also be recorded and included in the evaluation as an auxiliary criterion, especially for late orders. Figure 3 illustrates this: Although the earlier processing of sub-step *d* does not change the completion time of the order, it does make it possible to start sub-step *e* earlier and thus to complete the order earlier.

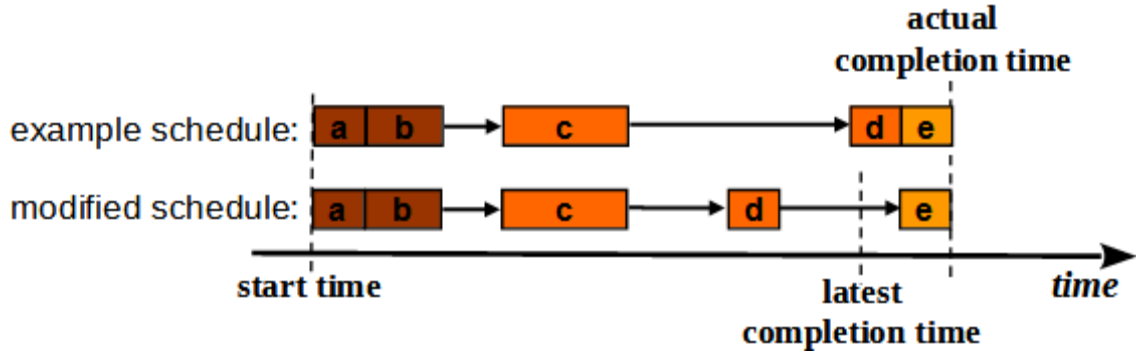

**Fig. 3:** Example of late completion of an order consisting of the five work steps *a - e*. The earlier processing of step *d* does not yet result in earlier completion and therefore does not improve the assessment of compliance with the deadlines. However, it is a necessary intermediate step to achieve this primary criterion and therefore earlier start of work-steps like step *d* must be evaluated positively as an auxiliary criterion.

### 4.3. Application Examples

Since the possibilities for designing an evaluation suitable for search have already been discussed in the previous sections and the topic will be addressed again in Section 4.4, this section is mainly concerned with aspects of coding and interpretation.

The different approaches to coding presented in the examples are intended to illustrate that there is no one solution, but rather a greater or lesser variety of design options. It is important to consider the coding that best serves a search, based on the circumstances of the task. A guideline can be the question, which coding corresponds rather to the causality principle, that is that small changes have also only small effects. The appropriate MH would then have to be selected. Further requirements for a suitable coding concern the coverage of the entire search space and the complexity of the interpretation, whereby the latter has rather subordinate importance in case of doubt.

In the first subsection, a number of alternative coding options for a rather simple combinatorial problem, the TSP, are discussed. In the use case presented in the second subsection, the impact of an encoding alternative on the options for repairing restriction violations during interpretation is considered. Finally, for the application presented in the third subsection, the discussion focuses around clever interpretation to avoid restrictions and the resulting consequences for coding.

### 4.3.1. Coding and Interpretation Exemplified by the TSP

As a first example for the design possibilities in coding and interpretation, the TSP may serve, in which a given set of cities is to be visited in any order and no city may be visited more than once. The shortest tour is sought. This NP-complete problem is considered a classical benchmark for combinatorial tasks in general and tour planning in particular.

For an EA that has genetic operators to change the gene order, the simplest way is to assign a number from 1 to *n* to each of the *n* cities and map it directly to the *n* genes of the chromosome. This number assignment remains unchanged. A gene order changed by mutations and crossover is directly interpreted as a tour and its length is calculated. Since each city occurs only once in the chromosome, the restriction of a single visit to each city is always satisfied.

But if the available EA leaves the gene order unchanged, one can, for example, encode the decision about the next city in the genes. Then the genes represent the index of the next city to be visited, where this index is related to a list of cities not yet visited. For the first gene, this list includes all *n* cities, for the second the remaining *n-1* cities, and so on. Accordingly, the value set of the genes decreases continuously. This procedure also avoids the multiple visit of a city. However, the interpretation effort is considerably higher, since new lists of the remaining cities have to be generated for each evaluation and each gene. Also, the EA must allow own value ranges per gene.

But what to do if it is a simple EA that limits the range of values uniformly for all DVs? In this case, a genetic repair can be considered, which, if the index is too large, simply dices it out again in the respective applicable range, whereby the new value is to be entered into the affected gene. The alternative idea of a phenotypic repair by reducing too large indices by modulo operation has a significant disadvantage: Thus, smaller indices occur more frequently after the adjustment than larger ones, which leads to a drift in the direction of the table beginning of the respective selection lists. However, mutations must be undirected, and this phenotypic repair gives them a quasi-subsequent favored direction. Therefore, genotypic repair is preferable in this case.

As another alternative possibility with an EA without sequence variations of the genes, one can use a chromosome consisting of up to *n* genes, which can take a value between 1 and *n-1*. The gene values are subject to mutation. At the beginning of the interpretation, the *n* cities are in a list in a fixed order from 1 to *n*. The genes are interpreted in the order pecified by the chromosome, with the *i*-th gene causing a shift of the city at the *i*-th position in the list by the gene value modulo *n*. The resulting city order in the list after processing the chromosome is the resulting tour, where again each city can occur only once. It should be noted that the minimum chromosome length should be *n/2*.

In this application, the first presented coding with an EA changing the gene order can be considered as the first choice.

### 4.3.2. Scheduling with Extended Start Time Planning

In Section 4.2, a scheduling task was discussed with the constraint that energy consumption peaks are to be minimized or better avoided at all. Such consumption peaks can be caused by energy intensive operations like heating, electric welding or shock freezing. Without this constraint, one of the common goals of scheduling tasks, namely a short total completion time, would lead to scheduling all work steps as early as possible, i.e. as soon as the first suitable resource (e.g. work station) becomes available. However, in order to smooth energy consumption, it may be necessary to delay the start of some energy-intensive work steps. Consequently, all such work steps must be scheduled with a start time. This can be specified as an absolute value or as a delay compared to the earliest possible start.

The decision on this alternative has, among others, effects on the remaining repair possibilities for such applications, where additionally also sequence restrictions of the processing steps have to be observed. The genotypic repair discussed in Section 4.1.4 is thus no longer possible with absolute start times, since the order of the genes in the chromosome has only limited relevance to the resulting processing order due to the absolute time specifications. And this the more, the more energy-intensive processing steps are to be planned. The gene order mainly influences the scheduling of the standard work steps and is reduced to the solution of possible allocation conflicts[4] for the others. The situation is different with phenotypical repair: Since for this the scheduling is postponed until all necessary preceding steps have been completed, the absolute start time specification is basically ignored. Ergo, absolute start time specifications and repair approaches are a poor fit.

If, on the other hand, one decides for delays in the start times, both types of repair are still possible. A genotypic repair moves the gene of the affected work step behind the last gene of the predecessor work steps and the phenotypic repair postpones its processing until all predecessor steps have been completed. In both cases, the delay stored in the gene is then added to the earliest possible start time.

Another consequence of the decision on the alternative described at the beginning concerns the character of the resulting task. When choosing start time delays, the combinatorial character is largely preserved. Absolute start times, on the other hand, reduce the importance of gene order the more steps and thus genes are affected. A combinatorial task thus increasingly becomes an integer optimization problem and the gene order only decides on the solution of allocation conflicts, i.e. it plays a lesser role than before.

The last question to be addressed is how to map the above gene model to chromosomes. As an example, an integer-coded EA is assumed, which does not provide for sequence changes of the genes. The gene model with start time delays is used, and similar procedures can be followed for the other variant. It should be emphasized that other coding alternatives are possible and reference is made to the literature. The chromosome is first divided into two parts, one determining the gene order and the second containing the DVs of those genes with start time delays. The interpretation is such that the genes of the first chromosome part determine the order of the work steps by swapping the steps in a list. In this process, the $i$-th gene interchanges the work step at the $i$-th position in the list with the list entry at position $(i+d)\,mod(n+1)$, where the distance $d$ is given by the gene and $n$ is the number of work steps. If the end of the list is exceeded, the modulo operation causes a continuation of the determination of the exchange partner at the beginning of the list.

Figure 4 shows a simple example consisting of genes for five work steps, three of which may have a start time delay. Let the work steps be labeled by *a* through *e*, and steps *b*, *d*, and *e* can be delayed. The genes for determining the sequence are located in the chromosome at positions 1 - 5 and those for determining the time delay at positions 6 - 8.

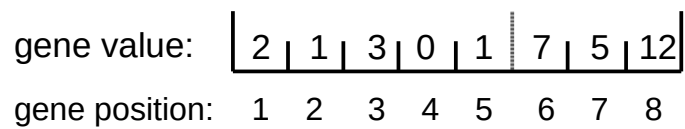

**Fig. 4:** Example chromosome for five work steps of which three can be started delayed

---

4   Allocation conflicts can arise when more than one gene claim a resource at the same time. Then, typically, the gene that is interpreted first, i.e. is closer to the beginning of the chromosome, is given preference.

Table 1 shows the stepwise change of the work step list when working through the first five genes of the example chromosome from Fig. 4. The row for gene 4 is omitted since it leaves the sequence unchanged.

| Processing of | Work Steps | | | | |
|---|---|---|---|---|---|
| initial list | a | b | c | d | e |
| gene 1 | c | b | a | d | e |
| gene 2 | c | a | b | d | e |
| gene 3 | b | a | c | d | e |
| gene 5 | e | a | c | d | b |
| result list | **e** | **a** | **c** | **d** | **b** |

**Table 1:** Determination of the work step sequence by permutations according to the gene values of the example chromosome from Fig. 4

In the last step, the delays are assigned: The value 7 of gene 6 to step b, the 5 of gene 7 to step d, and the 12 of gene 8 to step e. This assignment would leave the delays associated with the respective steps. Of course, one could also proceed differently and assign the delays to the work steps in their changed order, so that step e would be started 7 time units, d 5 and b 12 units later. Whether one of the two interpretation variants leads to better results, and if so which one, is probably application-dependent.

### 4.3.3. Layout Planning as an Example for Smart Handling of Complex Constraints

The last example is intended to illustrate how the search space can be meaningfully reduced by clever interpretation and how some restrictions can be taken care of at the same time. The following simplified task from the field of arrangement planning may serve this purpose: On a rectangular surface of width *w* and length *l* five different geometric types of objects are to be arranged in such a way that as little area as possible remains unused. These are triangular, square, rectangular, circular and oval objects of given dimensions, with the triangles being equilateral. The objects can be rotated and must not overlap after placement. Approximately the same number of all five types of objects are to be planned, whereby deviations of up to 10% are tolerable.

Two primary objectives arise from the task description: The unused area should be as small as possible, and the difference between the least and greatest number of placed objects per type should be no more than 10% of the least scheduled. In addition to the prohibited overlapping of objects, another restriction is that all objects must be located entirely on the surface.

The coordinates of the centers of the objects and a rotation can be considered as DVs. It makes sense to reduce the ranges of rotation angles in such a way that the isomorphisms of the geometric figures are exploited, which among others eliminates the rotation angle for the circular objects.

The approach that seems obvious at first glance is to let a MH determine the remaining rotation angles together with the center coordinates. However, this will lead to a large number of restriction violations due to overlaps and is therefore a possible approach, but not a clever one.

It makes more sense to replace the "free placement" by coordinate specifications with the following: In addition to the rotation, only the position along the width is determined by the MH and the rotated object is moved to the left in the direction of the length until it either arrives at the end or meets another object, see Figs. 5a and 5b. The value range of the coordinate $w$ may only be restricted in such a way that a fitting placement of the suitably rotated object to the respective edge remains possible, as shown in Fig. 5a below. For other rotation angles, this may result in edge protrusions. These are eliminated by a phenotypic repair in which the object is moved away from the edge as far as necessary, see Fig. 5a, upper rectangle and small green arrow indicating the repair. In addition, the MH determines the order of the objects to be placed in this way and thus ensures mixing on the surface, which can lead to favorable space utilization if placement order and rotations are appropriate. Overlaps and edge protrusions are thus avoided.

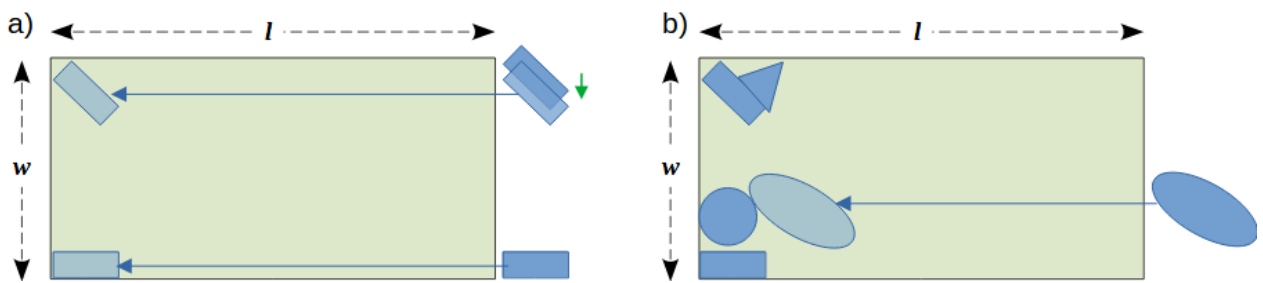

Fig. 5: Positioning of objects on a rectangular surface. On the left, a phenotypic repair (green arrow) is shown for the upper object. In the right part, the movement is stopped by contact with another object.

Since the number of plannable objects depends on the packing density and is thus not known a priori, a chromosome must contain an unknown but sufficient number of genes. This requires an estimation of the maximum conceivable objects on the surface. If an object could not be placed on the surface because it protrudes beyond the right edge of the surface despite a left shift, the gene is discarded[5]. Therefore, slightly more genes are to be foreseen than the estimation showed.

The interpretation ends when the chromosome ends or when the surface is filled. An indicator for a possible filling can be if an object could not be placed on the surface, for example, three times in succession because it protrudes beyond the right edge. Then it is checked whether the smallest object can still be placed with a suitable rotation. If not, the surface is considered filled. Otherwise, it should be checked whether the remaining space is sufficient for another smallest object. If yes, the surface is considered not filled and the chromosome is further interpreted. Otherwise, the placement of the first smallest object is valid and the surface is filled. Variants to this procedure can be considered, e.g. the termination due to reaching the end of the chromosome can be avoided by continuing the interpretation with the genes at the beginning of it, so that only the detection of the surface filling ends the interpretation.

Thus, in the described solution for coding and interpretation, the task contains a combinatorial part and an optimization of parameters (coordinate $w$ and rotation angle $α$ for all non-circular objects), where their reasonable value ranges depend on the object type and dimensions.

In the case of an EA, a problem-oriented mapping of DVs would result in genes that determine the object type and each have two or one DVs with object type-specific value ranges, namely $w$ and α, if applicable. The DVs of the genes and the order of the objects are subject to evolutionary change. If we want to apply the approach to coding of the previous example here, the following problems arise: While in the previous scheduling task the number of elements (work steps) to be scheduled was fixed, in this task it is not only variable overall, but moreover it is a priori unclear how many of which object type are to be placed. This makes the use of an EA based on fixed-length chromosomes consisting of integers or real numbers considerably more difficult, and reference is made to the literature on comparable applications.

---

5   Alternatively, a phenotypic repair in the form of an appropriate rotation of the object can be attempted.

However, there is at least one EA that offers a simple solution for the described coding task, namely GLEAM [6, 10, 15, 11]. In GLEAM the genes are typed[6] and have accordingly a gene type ID. Under this ID, the respective configurable number of integer and/or real parameters for the DVs is stored, for each of which a suitable range of values can be specified that is observed by the mutation operators. In the present example, there is one gene type per object type with the DVs constrained accordingly and a chromosome consists of any number of genes of the five gene types, see Fig. 6. The sequence is relevant to meaning and the number of genes of a chromosome is subject to evolution and would correspond to the estimated maximum size when generated. If chromosomes that are too short arise during evolution, the beginning of the chromosome can be used to continue the interpretation, as described previously. However, it is recommended to devalue too short chromosomes somewhat, e.g. according to the number of objects that could still be placed after reaching the end of the chromosome.

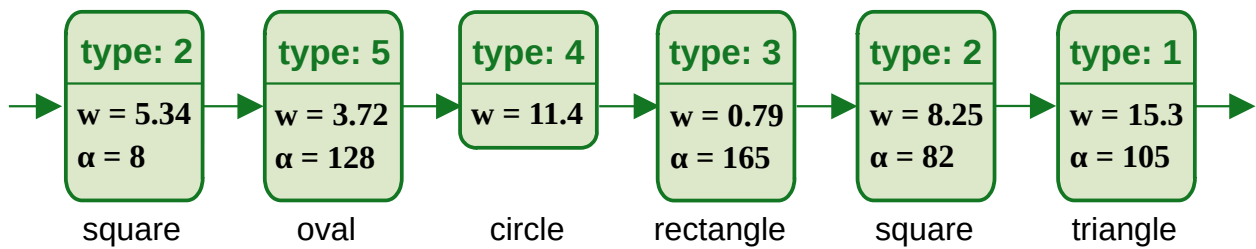

**Fig. 6:** Part of an example chromosome for the above gene model in GLEAM. Each gene contains the DVs assigned to the gene type as parameters and is an element of a linearly linked list. The list representation greatly facilitates gene sequence mutations. This includes the shifting of single genes as well as whole gene segments and the inversion of gene segments. In this example, the angles are restricted to integers, since it is assumed that this granularity is sufficient.

### 4.4. Pareto-Optimization or Fitness as Weighted Sum

Practical applications of MHs are usually multi-objective optimizations, i.e., they have multiple criteria to be optimized simultaneously. A common approach to computing the quality or fitness of a solution is to map the fulfillment of each criterion to a uniform quality scale, weight them, and add the weighted quality values to an overall quality or fitness. Restrictions can be treated as additional criteria, see Section 4.1.3, or alternatively as penalty functions, where the extent of a restriction violation is mapped to a scale from 0 (maximum violation) to 1 (no violation). The factor obtained in this way is then multiplied by the previously calculated *raw fitness* to obtain the final fitness. This can also be done with more than one penalty function. This procedure is known as *weighted sum* and the combination with penalty functions is a frequent extension, see [22]. Sometimes the criteria are also converted to costs and a cost function is created. In the end, however, this is also nothing else than a weighted sum, only that here money is taken as a quality measure.

---

[6] Therefore they have their own data type comparable to a `struct` of the programming language C or the data part of a class of the object-oriented languages.

Using the weighted sum to aggregate different criteria into one fitness value is relatively simple, but it has some disadvantages. One disadvantage is that weights must be assigned prior to optimization, i.e., at a time when one does not even know what is achievable and how difficult it is. Indeed, it may be useful and necessary to give higher weights to difficult-to-achieve objectives. Moreover, such aggregated criteria can compensate each other in an undesirable way. Measures against this are described in [22, Sect. 3]. In general, the exact determination of the normalization and possible penalty functions, the weights, and the allocation of the treatment of restrictions to additional criteria or penalty functions will be an iterative process.

The Pareto optimization [23, 24, 25] is an alternative to the weighted sum. It aims at finding good compromises, i.e. solutions where each of the criteria can be improved only by worsening at least one of the others. Therefore, the Pareto optimization returns a set of solutions that lie on a line called *Pareto front* and partially limit the set of permissible solutions $P$. Figure 7 shows an example for the case of two criteria to be maximized.

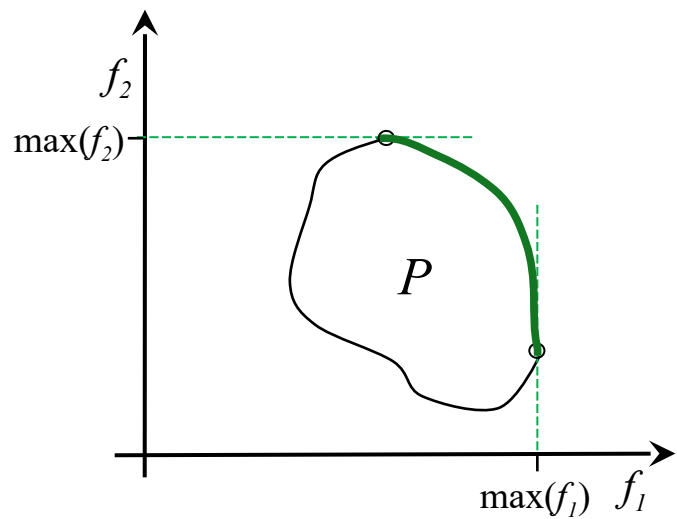

**Fig. 7:** Pareto front (green part) of two objectives to be maximized

With a suitable weighting, each point on a convex Pareto front can be approximated by the weighted sum, as shown in Fig. 8 on the left for the point S. If, on the other hand, the Pareto front has a non-convex part, this part is unreachable for optimization with the weighted sum. In Fig. 8 right, this would be the region between points A and B. A detailed discussion of this topic can be found in [22].

The advantage of Pareto optimization is that all points on the front are reachable regardless of their shape, that no weights have to be assigned a priori, and that it provides an overview of equivalent alternative solutions. The disadvantage is the increased effort of computing a large number of Pareto optimal solutions and the fact that presenting and evaluating their results from four criteria onwards becomes very difficult, to say the least. This issue is discussed in detail in [22] and [24]. Therefore, if there are more than three criteria, it is recommended to summarize part of them using the weighted sum.

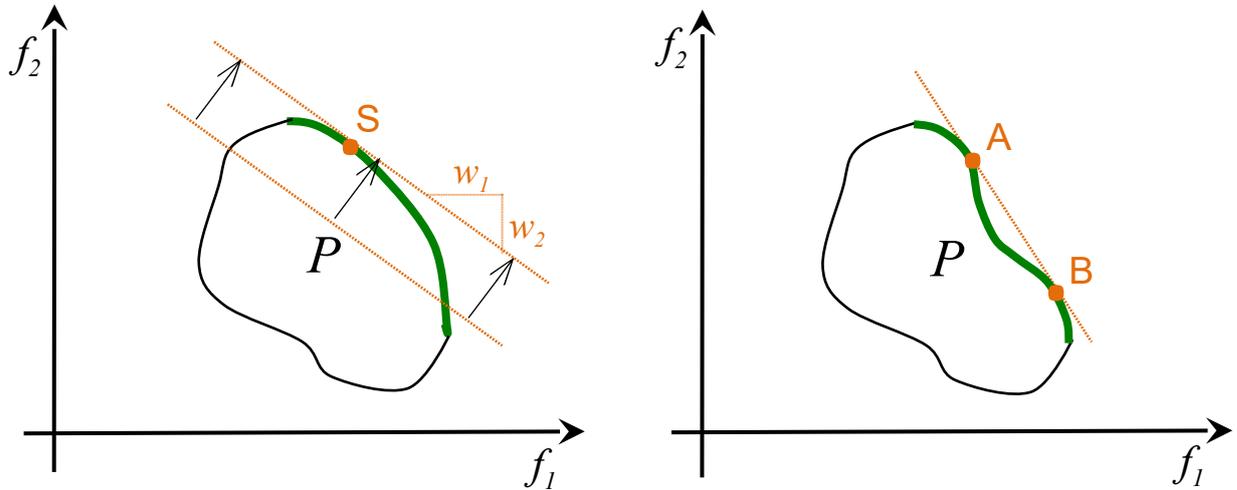

**Fig. 8:** The left side shows that a point *S* of the Pareto front can be found by suitable weights $w_1$ and $w_2$. The non-convex part of the front between points *A* and *B* in the right picture, on the other hand, cannot be reached.

The question of when to use which evaluation type advantageously is dealt with in detail in [22]. At this point, it should only be pointed out that this depends on the type and objective of the optimization project. For dealing with a new problem for which there is little prior knowledge, Pareto optimization with possible necessary aggregation of part of the criteria is meaningful. If, on the other hand, similar tasks from the same field of application are to be processed again and again, as may be the case, for example, in job shop scheduling, then, after a previous analysis of the alternatives by means of Pareto optimization, it is in many cases no longer of interest to determine the entire Pareto front each time and to have a human decision maker select a solution from it. Rather, one will be interested in a solution with a corresponding weighting of the alternatives, which is obtained in an automated process. In such cases, the weighted sum will be purposeful with less effort.

Special EAs for the determination of the Pareto front have been developed, which can determine the front in one run if the population is sufficiently large. To these belong the two algorithms NSGA II [26] and NSGA III [27, 28], which can be considered as standard procedures, and the SPEA2 [29].

## 5. Memetic Extension of EAs

Among the strengths of EAs is the global nature of their search and among their weaknesses the poor convergence properties to an optimum. With local search methods and local hill climbers it is just the other way round, many show good convergence properties, but ignore better (local) optima in the vicinity, if a valley would have to be crossed to reach them, i.e. a decrease in quality would have to be accepted. The obvious idea is to combine both methods in order to exploit the respective advantages.

There are two approaches to this that have proven successful and can be used in a complementary manner:

- When forming the initial population, some or all of the randomly generated individuals are locally enhanced, or heuristics are used to generate some initial individuals. In the latter case, only a small fraction (e.g., 20% at most) should be "pre-generated" to ensure sufficient search space coverage of the initial population. Also, heuristics tend to generate more or less similar solutions.

- All or part of the offspring generated by the genetic operators are locally improved. This means that the local search is parallel to the global one and, from the EA's point of view, virtually only the mountain tops matter. This form is also called Memetic Algorithms (MA) [12 - 14] and is the basis for an own sub-discipline of EAs, see e.g. [30].

The decisive question is: Is the effort for the local search worth it? It could be shown that especially in the area of continuous and mixed-integer optimization considerable performance increases measured in the number of calls to the fitness function can be achieved [13, 14, 31]. Depending on the application, reductions in the number of average fitness calculations on the order of factors up to $10^3$ were observed [14, 32]. Another advantage is that the range of favorable population sizes (cf. Section 8) with an MA is significantly smaller than when using the associated base EA [14]. It should be emphasized, however, that an unfavorable choice an LS or a heuristic can also lead to failures [13, 14, 31, 33].

In general, the supplementation of the evolutionary search by suitable local methods or also application-related heuristics is also theoretically justified by the no-free-lunch theorems [34, 35]. In summary, these theorems state that, with respect to the set of all mathematically possible problems, all search algorithms are on average equally good (or equally bad). Conversely, this means that there is no universal algorithm that solves all optimization problems most efficiently. So it makes sense to integrate application-related knowledge into an optimization procedure, be it through initial solutions generated otherwise or through LS or heuristics that serve as memes of an MA.

An MA introduces a number of design issues that are either hard-coded or lead to further strategy parameters:

1. Which local searcher (LS) should be used?
   Should it be simple, imprecise and fast or more elaborate, precise and slow (i.e. require more fitness calculations)? Should it be general or application-specific? The choice of a suited LS can be of crucial importance to success, see e.g. [13, 31, 32, 33].

2. Should the LS result be used to update the chromosome or not?
   This issue is discussed controversially in literature and depends on other measures taken to prevent premature convergence, like using a structured population [36] instead of a panmictic one. This question is discussed in more detail in [14].

3. How are offspring selected to undergo local search?
   By chance or according to fitness? See also [13].

4. How often should local improvement be applied or to which fraction of the generated offspring per generation?

5. How long should the local search be performed and how precise should it be as a result?

The last two questions determine the proportion between local and global search. The answer to most of the above questions will be application-dependent, prohibiting hard coding. However, since additional strategy parameters also increase the possibility for inappropriate settings, cost-benefit-based adaptation was experimented with early on [31, 37]. This allows adaptive selection of an LS from a given set as well as favorable adjustment of the parameters associated with design questions 4 and 5 [14].

## 6. Remarks on Comparison between Metaheuristics and their Variants

A comparison between alternative MHs, different settings of their strategy parameters such as population size (see also Section 8), the efficiency of different LHCs in a memetic algorithm, or the like cannot be based on a comparison of single runs because of the stochastic nature of the methods. Rather, multiple runs comparing the means or medians of the measured quantities are required. If the differences are small or the measured values are (highly) scattered, statistical evaluation methods are required. When comparing two alternatives, the calculation of confidence intervals and a simple *t*-test may be sufficient; when there are more alternatives to be compared, an analysis of variance (see e.g. [38]) is indicated.

There are two different measurement methods for a comparison: Either a certain target quality is specified and the number of evaluations required to achieve it is counted, or the quality achieved is measured for a specified number of evaluations. The latter approach is more manageable in terms of effort and, with a suitable limit on the number of individuals evaluated, usually yields the more significant results. If the application allows, at least 30 or better 50 runs should be performed per procedure setting to be compared. If this value has to be reduced for reasons of effort, the significance of the results decreases as the number of runs is reduced. Less than 10 runs are not recommended.

When comparing different MHs, ensure that the application-dependent strategy parameters are appropriately set for each MH separately before comparison. In any case, for EAs and MAs, the population size should be appropriately selected as described in Section 8.

## 7. Search Reliability and Population Structures

MHs like EAs cannot in principle guarantee that the global optimum will be found in finite time. Since the optimum is usually not known in practical applications, there is always some uncertainty as to whether the present result may not be surpassed. The algorithm is usually allowed to run until either a given time is used up or a predefined number of generations or quality has been reached. However, it is better to measure the stagnation and to stop, if e.g. no fitness progress can be observed after *n* successive generations. Then the population is considered as *converged*. We will come back to this at the end of this section.

Populations have been shown, however, to converge quite rapidly and before reaching the global optimum or at least a good sub-optimum if a (slightly) better offspring can spread undisturbed over several generations. This may be comparatively common in unstructured populations, where any individual can in principle produce offspring with any other. In such a case, depth search becomes too predominant over breadth search. Such populations are called *panmictic pop-* *ulations* and they do not correspond to the biological paragon of EAs. In nature, a population of basically reproductive individuals is relatively separated by spatial distance. Therefore, early in the evolution of EAs, structured populations were experimented with and shown to effectively counteract premature convergence, making them superior to unstructured populations in terms of achieving good to optimal solutions [36, 39]. Moreover, they scatter less from run to run [40, 41]. This has been experimentally verified not only for GAs and the evolution strategy [36, 39 - 42] but also corresponds to the author's long experience with the various GLEAM and HyGLEAM applications [10, 14, 15].

The following two principally different types of structuring can be distinguished: First, the separation of a population into sub populations (islands) that evolve separately over a longer period of time and exchange individuals only from time to time, and second, the division into overlapping neighborhoods. Island models have the disadvantage of adding a larger number of new strategy parameters: size and number of islands as well as their connections and the control of exchange: how many and which individuals migrate to another island and who is replaced there? When does exchange occur: statically every *n* generations or depending on local stagnation? Populations structured according to the island model are thus much more difficult to handle and will therefore not be considered further here.

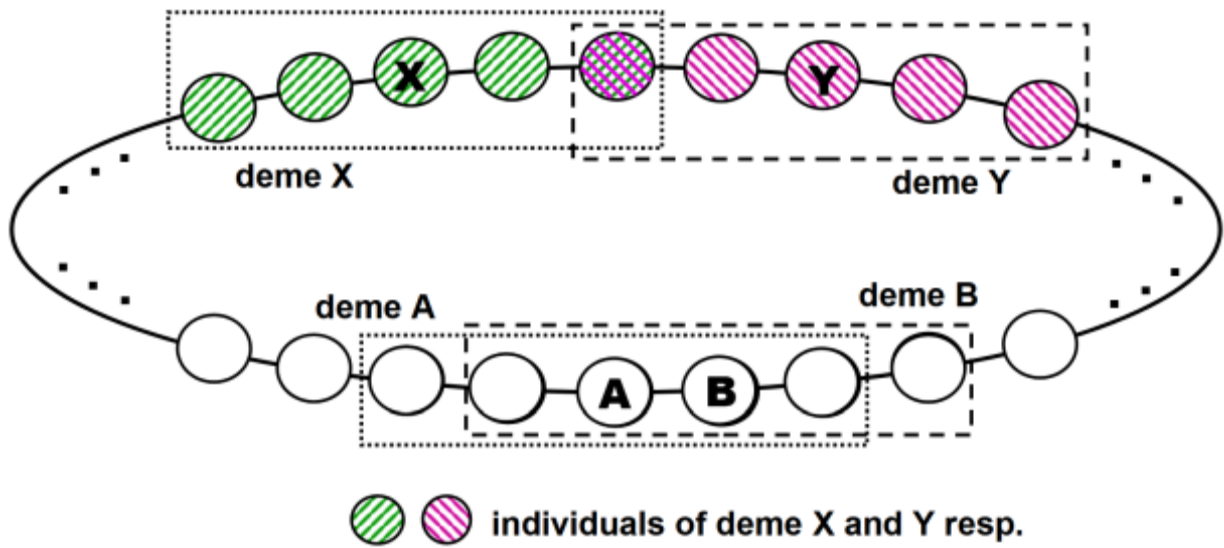

**Fig. 9:** Diffusion model based on overlapping neighborhoods, called demes. The demes of the two individuals "X" and "Y" have a minimum overlap, while "A" and "B" have a maximum one. Typical deme sizes range from 7 to 11, depending on population size.

In the alternative of overlapping neighborhoods, the individuals are arranged, for example, on a ring that implies a geographic neighborhood, regardless of their phenotypic characteristics. The exemplary neighborhood of individual "X" shown in Fig. 9 consists of the two individuals to the right and to the left. Together with "X" they form the so-called *deme* of "X". Each deme represents a panmictic sub population within which mates are selected and the acceptance of offspring by replacing the parent occurs. The rules for offspring acceptance are based on the neighborhood: for example, it may be specified that the best offspring must be better than the parent being replaced or, less strictly, only better than the worst individual in the deme. Because demes overlap, as shown in Fig. 9, genotypic information can spread across neighborhood boundaries. Therefore, they are also referred to as *diffusion models*. Because this spread is much slower than in panmic-tic populations, niches of more or less similar individuals can emerge, evolve, spread, clash, and compete. This maintains genotypic diversity over a longer period of time. In addition, diffusion models such as this induce an adaptive balance between breadth and depth search. Further details and alternatives to the ring structure can be found in [10, 36, 39, 40, 42].

Regarding the organization of the population, the following is recommended when selecting an EA or MA:

- When runtime is limited and fast solutions are important, with suboptimality being acceptable, one can work with panmictic populations. The population size should be sufficiently large to avoid too fast premature convergence.

- If one aims at a good trade-off between runtime and solution quality, diffusion models based on a two-dimensional mesh are a good approach. Such neighborhoods are also known as cellular EAs or MAs [39, 40, 42].

- If, on the other hand, the goal is to achieve the best possible quality while accepting longer runtimes, then ring-based neighborhoods are the first choice [10, 14, 40].

The neighborhood model also allows a more sophisticated definition of two stagnation indicators that can serve as termination criteria for a run. In each case, the generations are counted for which, in succession

- no improvement of the best individual of each deme occurs

- there is no acceptance of a descendant per deme.

As a rule, the improvement of the demes will stop at first. At the latest when there is no more acceptance for several generations, the population can be considered as converged. Meaningful limit values are application-dependent, whereby it can be assumed that with the duration of missing acceptance the probability drops that it comes nevertheless still to an accepted or even better descendant.

## 8. Handling Strategy Parameters and the Population Size in Particular

In general, it is easier for a novice to choose a MH that requires the setting of as few strategy parameters as possible. It is also strongly advised to consult the literature for experience reports on the selected MH, if possible applied to comparable tasks, and to take the settings reported there as a basis. When experimenting with strategy parameters, one is reminded of the hints given in Section 6.

For population-based MHs, however, one parameter will always have to be adjusted to the current task, namely the population size $\mu$. For structured populations, this should be at least twice as large as the deme size, although significantly more is certainly better. A favorable population size depends not only on the application but can also be determined by the chosen coding and interpretation. For example, if all decision variables are encoded in the chromosome, this results in a different search space than if, for example, a part is determined by heuristics during interpretation, see Section 4.3.3 and [15].

To explain the general approach for determining a suitable population size $\mu$, we draw on experience with benchmark functions whose optimum is known. Fig. 10 shows the effort measured in fitness calculations (blue line) necessary to achieve a given target fitness for different population sizes. For very small values of $\mu$, an otherwise high effort (dashed line) may have been limited by another termination criterion such as stagnation, see Section 7. If $\mu$ is too low, more or less runs do not reach the target fitness and are therefore considered unsuccessful. As $\mu$ increases, success occurs more frequently until finally all runs are successful (working area). If $\mu$ is increased further, the effort also increases without being necessary or useful.

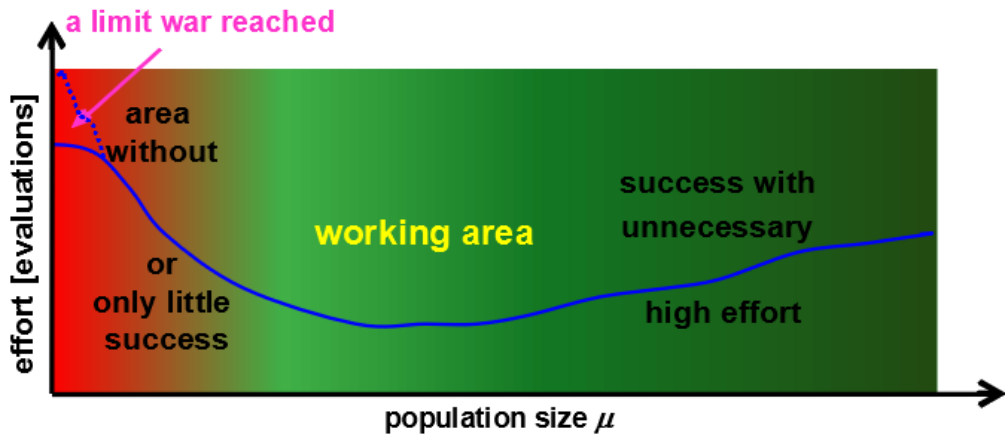

**Fig. 10:** Relationship between population size $\mu$ and success as well as effort

For a new application, the attainable fitness is usually unknown and it must be determined experimentally what should be considered a good solution and thus a success. The determination of a suitable starting value for $\mu$ can be seen as a more or less well estimated guessing and it is recommended to begin with a rather to the large $\mu$. For an EA, values of some 100 individuals are nothing unusual, whereas for an MA based on this EA, much lower values are sufficient, e.g. $\mu/8$ or $\mu/10$ of a good value of $\mu$ for the EA involved, see also Section 5. The determination of a suitable target fitness is begun with the starting value for $\mu$ thus determined with at least 10 runs. If these runs all yield similar fitness values with similar effort, the result should be secured with 10 or better 20 more runs. Such a result also shows that the target quality for the found population size is stably reached and we are somewhere in the green zone. Otherwise, $\mu$ must be increased until such a result is obtained. Based on the runs for the population size found, two termination criteria are now determined for the further runs: The average of the achieved fitness values gives the target fitness, and the maximum of the required fitness calculations plus an additional amount is taken as the effort limit. Runs that exceed this limit without reaching the target fitness are considered unsuccessful. Depending on the time spent, the surcharge can be small (e.g. 20%) or larger (50-100% of the calculated effort maximum).

In order to determine the limits of the working area, $\mu$ is reduced step by step as long as the target fitness is reliably achieved with decreasing effort. As soon as the values for the effort begin to scatter more strongly, we approach the left limit of the working range. In Fig.10, this is the light green area. When the first non-successful runs occur, we are in the red/green area and the population size is too small. The size we are looking for is then in the area to the right of this and we should choose it so that the effort does not scatter too much.

## 9.  Search Speed, Archive and Parallelization

As mentioned in Section 2, EAs require the evaluation of a large number of alternative solutions, which can be assumed to be in the tens of thousands or more, depending on the application. For linguistic simplicity, the steps involved in evaluating a solution represented by an individual, which may require simulation or other calculations of the evaluation criteria, are collectively referred to as evaluation. The result is the values of the assessment criteria of an individual. This would then be followed by either calculating the weighted sum or determining the dominance properties in Pareto optimization.

EAs, like other population-based MHs, are inherently parallel. Individuals evolve in parallel and independently, except for mating and crossover of offspring production. Thus, it is natural to distribute a population across multiple computers. Appropriate approaches have also existed at an early stage [36, 43, 44, 10]. On the other hand, in almost all application projects it can be assumed that the evaluation takes much longer than the creation and manipulations of the chromosome from which the solution to be evaluated has emerged. Thus, the question arises, should one parallelize the population or distribute the evaluations?

When parallelizing the population, the underlying population model plays an important role. In the neighborhood model (see also segment 7), a high degree of parallelization is possible up to a one-to-one assignment of individuals to computer nodes, whereby the neighbors of the demes of the individual(s) of a computer node must be available to this node. In the early stages of evolution, this leads to frequent updates and a correspondingly high but local communication overhead. Another positive property is the lack of a central coordinating instance as would be required in a panmictic population. Thus, parallelization according to the neighborhood model is especially suitable for parallel computers that support local communication, see also [36, 43, 45]. The situation is different, however, for parallelization according to the island model, where the communication overhead is significantly lower if each island is managed by one computer node [44, 45, 46]. However, this limits the possible degree of parallelization. If, on the other hand, the evaluations are distributed, one has to deal with a mostly constant communication overhead and constant computational load, provided that the computational time required to evaluate a solution is independent of the solution itself. Both parallelization approaches can, of course, be advantageously combined [47]. Modern parallelization and virtualization techniques such as microservices, container virtualization and the publish/subscribe messaging paradigm allow flexible and scalable implementations that, due to their generic approach, enable easy interchangeability of the software used, especially for evaluation [46, 47, 48]. Since evaluation is always application-dependent, this application-specific component of metaheuristic optimization can thus be replaced more easily than in most other implementations.

In addition to these basic and diverse possibilities for increasing the practicable field of application for EAs through parallelization, there is also the use of a solution archive. The idea behind this is that in evolutionary search it can happen that a solution is generated multiple times, so that there are multiple tests of the same point in the search space. Or that two solutions are sufficiently similar in terms of practical realization so that they can be considered equivalent. This can occur especially in the optimization of continuous decision variables. For example, the dimensions of work pieces may be mathematically different even if these differences cannot be represented in a practical production and would not matter. Such similarity measures are of course to be determined application-dependently and they can serve as a basis for tracing similar solution proposals back to a once evaluated individual. Thus, in the course of optimization, an archive of representative solutions is created, which can be referred to before performing an evaluation.

## 10. Some Aspects of Project Management

Engineers tend to consider technical tasks in a project more important than interpersonal problems of the people involved in the project. Therefore, it is expressly warned against underestimating the possible negative consequences of group dynamic processes and possible animosities or diverging interests of the project participants. In the following, two scenarios are examined in more detail: An optimization project to improve existing processes or products and aspects of presenting the results of an optimization project.

If existing products or processes such as production planning are to be optimized, there are usually already employees in the company or organization who are involved in this and who usually enjoy a high reputation for their difficult high-quality work. Intervening here as an external service provider or new specialist department can be perceived as an attack on the position and disempowerment of the previous expert department. Also at least a part of the (middle) management will be skeptical of new procedures designed to replace established ones. Especially since these new methods are based on randomness and cannot even ensure to always provide the optimal solution at all times. The fact that the methods used up to now usually could not do this either is often conveniently overlooked.

It is therefore recommended to involve the affected technical department as far as possible. If you work as an optimizer against the previous experts, they might tend to look for relevant information that they can withhold inconspicuously in order to let the project fail and to maintain their status. An important counter-argument can be to make the results of previous planning or product design the basis for optimization with the MH. For this the data of previous results must be transferred suitably as start individuals or start solutions. This additional effort can be justified by the fact that now the optimization results can only become better than the previous solutions. This creates security and confidence, makes it more difficult for critics and involves the representatives of the previous approach in the project. A success of the optimization project is now also their success, because their previous work is now a valuable basis for the new procedure. Apart from that, it is also reasonable not to start a metaheuristic search from scratch, but to take previous or qualitatively inferior solutions as starting points, see also Section 5.

In many optimization projects, the results serve as the basis for decisions on how to proceed. In such cases, the presentation of the results in the form of a Pareto front is not only descriptive, but it also shows the possible decision alternatives that are equivalent in terms of the evaluation criteria, see also Section 4.4. Depending on the audience and the presentation possibilities, one has to limit oneself to the comparison of two target criteria or - provided that the projection technique is suitable - one can present the Pareto plane of three criteria.

This approach has several advantages: First, it involves the client or the higher-level management in the decision-making process; second, it delegates part of the responsibility upwards from the optimization team; and third, the decision-makers will identify much more with the project and defend it if necessary than if they had been presented with a finished result. After all, they have participated in the discussion, subjectively understood the alternatives, and finally decided.